\documentstyle[twocolumn,times]{article}

\newif\ifshorten    
\shortentrue

\newcommand{\RULE}[2]		                
{\begin{tabular}{c}
 \( {#1} \) \\ \hline \( {#2} \)
\end{tabular}}

\newcommand{\comment}[1]{}

\newtheorem{defi}{Definition}
\newtheorem{LEMMA}{Lemma}
\newtheorem{COROLLARY}{Corollary}
\newtheorem{thm}{Theorem}

\newcommand{\kases}{    
   \vspace{-3.2mm}
   \begin{list}
 {???}
 {\setlength{\leftmargin}{4.1mm}
  \setlength{\labelwidth}{2mm}}
  \setlength{\parsep}{1mm}
  \setlength{\itemsep}{0.01mm}
  \setlength{\topsep}{0.01mm}
  \setlength{\parskip}{0mm}
  \setlength{\parindent}{0mm}}
\newcommand{\Endkases}{\end{list}\vspace{-3.2mm}}

\newcommand{\etal}{{\em et al.}}

\newcommand{\State}[5]
{\mbox{\vbox{state #1:\\
\parbox[t]{2em}{\verb+ + B:} \parbox[t]{\ExampleWidth}{#2}\\
\parbox[t]{2em}{\verb+ + S:} \parbox[t]{\ExampleWidth}{#3}\\
\parbox[t]{2em}{\verb+ + I:} \parbox[t]{\ExampleWidth}{#4}\\
\parbox[t]{2em}{\verb+ + P:} \parbox[t]{\ExampleWidth}{#5}
}}}
\newcommand{\mytable}[1]{

\vspace{1mm}
\centerline{ #1 }

}
\newcommand{\th}[1]{$^{\mbox{#1}}$}

\newtheorem{example}{}    

\newlength{\ExampleWidth}
\newcommand{\startx} 
 {\begin{example}
  \rm
  \ \ \
  \begin{minipage}[t]{\ExampleWidth}}

\newcommand{\startxl}[1] 
 {\begin{example}
  \label{#1}
  \ \ \
  \rm
  \begin{minipage}[t]{\ExampleWidth}}

\newcommand{\stopx} 
 {\end{minipage}
  \end{example}}

\newcommand{\BetweenExamples}{}

\newlength{\ExampleWidthM}
\renewcommand{\startxl}[1]{
\begin{example}%
\label{#1}
   \
   \rm
   \begin{minipage}[t]{\ExampleWidth}}

\newcommand{\myitem}[2]{\mbox{{#1} \
		\begin{minipage}[t]{\ExampleWidthM}{#2}\end{minipage}}

\vspace{1.4mm}

}

\makeatletter
\def\@normalsize{\@setsize\normalsize{12pt}\xpt\@xpt
\abovedisplayskip 10pt plus2pt minus5pt\belowdisplayskip \abovedisplayskip
\abovedisplayshortskip \z@ plus3pt\belowdisplayshortskip 6pt plus3pt
minus3pt\let\@listi\@listI}

\def\subsize{\@setsize\subsize{12pt}\xipt\@xipt}

\newcommand{\mysection}[1]{

\vspace{-4.5mm}

\section{#1}

\vspace{-2.4mm}

\noindent}

\def\section{\@startsection {section}{1}{\z@}{24pt}
{12pt}{\center\large\bf}}

\def\subsection{\noindent \@startsection {subsection}{2}{\z@}{24pt}
{12pt}{\subsize\bf\hspace*{-0.3in}}}
\makeatother

\begin{document}
\ifshorten \date{} \fi

\title{\Large\bf \vspace*{0.25in} Resolution of Syntactic Ambiguity:
the Case of New Subjects\thanks{This paper appears in the Proceedings of the
Fifteenth Annual Meeting of the Cognitive Science Society (1993).  It is
electronically archived in the Computation and Language E-Print Archive as
cmp-lg/9406028.  I wish to thank Bob Frank, Susan Garnsey, Young-Suk Lee and
Mark Steedman for their helpful suggestions.  Any remaining errors are my own.
This research was supported by the following grants: {\sc DARPA}
N00014-90-J-1863, ARO DAAL03-89-C-0031, NSF IRI 90-16592, Ben Franklin
91S.3078C-1. }}

\author{{\bf Michael Niv} \\
Department of Computer and Information Science \\
University of Pennsylvania\\
200 South 33\th{rd} Street\\
Philadelphia, PA \ 19104\\
{\tt niv@linc.cis.upenn.edu}}

\maketitle
\thispagestyle{empty}

\mysection{Abstract}%
I review evidence for the claim that syntactic ambiguities are resolved on the
basis of the meaning of the competing analyses, not their structure.  I
identify a collection of ambiguities that do not yet have a meaning-based
account and propose one which is based on the interaction of discourse and
grammatical function.  I provide evidence for my proposal by examining
statistical properties of the Penn Treebank of syntactically annotated text.

\setcounter{page}{1}
\mysection{Introduction}%
On what basis do people resolve the syntactic ambiguity so common in language?
Some researchers (e.g. Frazier and Fodor 1978\nocite{sausage}, Mitchell Corley
and Garnham 1992\nocite{MCG92} and references therein) have argued that the
sentence processor embodies structurally defined criteria such as Minimal
Attachment and Late Closure.  Others (e.g.  Crain and Steedman
1985\nocite{CrainSteedman85}, Altmann and Steedman
1988\nocite{AltmannSteedman88}, Trueswell and Tanenhaus
1991\nocite{TrueswellTanenhaus91}) have argued that ambiguity resolution
decisions are made online, rather quickly, based on the relative sensibleness
of the available analyses.  Psycholinguistic research of this issue has
focused on a limited collection of structures such as:

\startxl{EG1}
\myitem{a.}{The horse raced past the barn fell.}
\myitem{b.}
	{The doctor told the patient that he was having trouble with to leave.}
\stopx

Out of context, sentences with structures similar to either of these examples
are garden-paths.  When such sentences are put in contexts that support the
correct reading, the garden-path effect disappears.  Some questions remain
about how quickly such discourse-sensitive sensibleness preferences are
brought to bear on the ambiguity resolution process (Mitchell \etal\
1992). But this issue will not be addressed further here.

One attractive aspect of structural preference theories is that two very
simple structural strategies can account for so much data.  Aside from the two
examples in \ref{EG1} Minimal Attachment defined in \ref{MAdef} predicts that
\ref{MA2} is a garden-path.

\startxl{MAdef}
Minimal Attachment  (Frazier and Fodor 1978):\\
Each lexical item (or other node) is to be attached into the
phrase marker [in the way which requires the smallest] possible number of
nonterminal nodes linking it with the nodes which are already present.
\stopx

\BetweenExamples

\startxl{MA2}
John has heard the joke about the pygmy is offensive.
\stopx

Late Closure \ref{LCdef} predicts difficulties with the sentences in \ref{LC1}.

\startxl{LCdef}
Late Closure (Frazier and Rayner 1982):\nocite{FrazierRayner82}
When possible, attach
incoming lexical items into the clause or phrase currently
being processed (i.e. the lowest
possible nonterminal node dominating the last item analyzed).
\stopx

\BetweenExamples

\startxl{LC1}
\myitem{a.}{John said that Bill will leave yesterday.}
\myitem{b.}{When the cannibals ate the missionaries drank.}
\myitem{c.}{Without her contributions failed to come in.
(from Pritchett 1988\nocite{Pritchett88})}
\myitem{d.}{When they were on the verge of winning the war against
Churchill and Roosevelt met in Yalta to divide up postwar Europe. (from Ladd
1992\nocite{Ladd29})}
\stopx

The claim of this paper is that ambiguity resolution decisions are based
solely on the sensibleness of the available readings.  I argue that for each
structure in \ref{EG1}, \ref{MA2}, and \ref{LC1}, aspects of meaning which
must be assumed for independent reasons are responsible for the ambiguity
resolution behavior observed in humans.

\mysection{Existing Accounts}%
Crain and Steedman (1985) argued that resolution of the ambiguity in
\ref{EG1}b is sensitive to the prior discourse.  When there is a unique
patient in the discourse, the NP `the patient' uniquely identifies him/her and
there is no need for further restrictive modifiers.  Consistent with Grice's
(1965)\nocite{Grice65} maxim of Manner (be brief), the hearer selects the
complement clause analysis of `that he$\ldots$'.  When there are two patients
in the discourse, the NP `the patient' does not pick out a unique referent, so
Grice's maxim of Quantity (make your contribution as informative as required)
guides the hearer to construe the continuation `that he$\ldots$' as a
restrictive relative clause.

When the sentence is presented out of context, the hearer must {\em
accommodate\/} a situation in which it is felicitous.  That is, the hearer
must change his/her mental model to support the presuppositions carried by the
sentence.  The restrictive relative clause reading requires accommodating a
more complex situation in which there is more than one patient, so according
to a preference for parsimony, the complement clause analysis is preferred.

In another study Crain and Steedman (1985) found that the garden path effect
in sentences with structures as in \ref{EG1}a is reduced/eliminated when the
main-verb reading is implausible.  This finding was replicated by Pearlmutter
and MacDonald (1992)\nocite{Pearlmutter92}.  Others (e.g. Trueswell and
Tanenhaus 1991) have found other aspects of meaning (e.g. temporal coherence)
are also relevant for the ambiguity in
\ref{EG1}a.

Niv (1992)\nocite{Niv92} argued that the tendency for low attachment of
adverbials such as `yesterday', as in \ref{LC1}a, results from a general
principle of competence to order constituents in the sentence in increasing
order of ``information volume''.  Attaching `yesterday' high is dispreferred
as this single-word constituent would have to follow the heavier constituent
`that Bill will leave'.  This preference disappears when the adverbial is made
heavier, e.g. `because he became very angry at us.'

Stowe (1989, experiment 1)\nocite{Stowe89} found that for a certain class of
ambiguous verbs, plausibility manipulations can affect the garden path effect
in \ref{LC1}b.  For verbs which exhibit causative/ergative alternations,
illustrated in \ref{CausativeErgative}, reducing the plausibility of the
subject to serve as the causal agent (e.g. \ref{Stowem}b) picks out the
ergative reading and avoids the garden path.

\startxl{CausativeErgative}
\begin{tabbing}
Causative: \= John moved the pencil.\\ Ergative: \> The pencil moved.
\end{tabbing}
\stopx

\BetweenExamples

\startxl{Stowem}
\myitem{a.}{Before the police stopped the driver was already getting nervous.}
\myitem{b.}{Before the truck stopped the driver was already getting nervous}
\stopx

Recent research on the local ambiguity in \ref{MA2} has focused on whether
subcategorization preferences of the matrix verb affect the garden-path.  When
the matrix verb is one which tends to take a clausal complement more often
than an NP complement, the garden path may be avoided.  As with other
psycholinguistic facts summarized here, the claims are still controversial,
but a recent experiment by Garnsey, Lotocky and McConkie (1992)\nocite{GLM92}
seems to settle the issue in favor of lexical preference effects.

In summary, there are meaning-based accounts for people's ambiguity resolution
preferences for \ref{EG1}a, b, and \ref{LC1}a.  Lexical preference accounts
for some of the observation, in \ref{MA2} and \ref{LC1}b, but cannot alone
account for the garden paths in
\ref{Remaining}.

\startxl{Remaining}
\myitem{a.}{John finally realized just how wrong he had been remained to be
 seen.}
\myitem{b.}{When Mary returned some of the presents were missing.}
\stopx

\ifshorten
 \noindent \ref{Remaining}a has the same structure as \ref{MA2}
 and the verb `realize' is
 biased toward a sentential complement reading (according to Garnsey \etal's
 findings, as well
 as the Brown Corpus).  But there is still a perceptible garden-path effect.
 \ref{Remaining}b has the same structure as \ref{LC1}b and the verb  `return'
 occurs more
 frequently in the Brown Corpus without an object than with one. But again,
 there is still a garden
 path effect.
\fi

\mysection{Avoid Subjects}%
All of the examples above that are not fully accounted for, i.e. \ref{LC1}c,
d, \ref{Remaining}a, b, exhibit a common property --- a certain NP, which I
will call the {\em critical NP,} has both a subject and a non-subject
analysis.  In each of these examples, the non-subject analysis is preferred.

There are a few hints in the literature that readers really do prefer to avoid
subjects.  One hint comes from a second experiment that Stowe (1989)
conducted.  In addition to manipulating the agenthood of the subject of the
first clause, this experiment also manipulated the plausibility of the
critical NP to serve as the direct object.  Sample experimental materials are
given in \ref{Stowem2}.

\startxl{Stowem2}
\begin{tabbing}
\ \ \ Implausible: \= \kill
Animate: \\
\ \ \ Plausible: \ \> When the police stopped the driver \\
                   \> became very frightened.\\
\ \ \ Implausible: \> When the police stopped the silence \\
                   \> became very frightening.\\
Inanimate:\\
\ \ \ Plausible:   \> When the truck stopped the driver \\
                   \>  became very frightened.\\
\ \ \ Implausible: \> When the truck stopped the silence \\
                   \> became very frightening.
\end{tabbing}
\stopx

\noindent  Stowe found implausibility effects at the critical NP even in
the inanimate sentences, where the readers exhibit commitment to the ergative
(intransitive) analysis.

Another hint comes from an experiment by Holmes, Kennedy and Murray
(1987)\nocite{HKM87}.  Using experimental materials as in \ref{HKM87m}, Holmes
\etal\ found that in the disambiguation region (either `to the officer' or
`had been changed') the transitive verb sentence (TR) was read substantially
faster than the other two sentences.

\startxl{HKM87m}
\addtolength{\ExampleWidthM}{-4mm}
\myitem{(TR)}
 {The maid disclosed the safe's location within the house to the officer.}
\myitem{(TC)}
 {The maid disclosed that the safe's location within the house had been
	changed.}
\myitem{(RC)}
	{The maid disclosed the safe's location within the house had
	been changed.}
\addtolength{\ExampleWidthM}{4mm}
\stopx

\noindent The that-complement (TC) sentence was read slightly faster than the
reduced complement (RC) sentence.  This finding was subsequently replicated by
Kennedy \etal\ (1989)\nocite{KMJR89}.  It is quite surprising that the
unambiguous condition TC should take longer than the locally ambiguous
condition TR.  Holmes \etal's speculation that beginning a new clause requires
additional processing is consistent with a strategy of avoiding analyzing NPs
as subjects.

My claim is not that the processor is averse to subjects in general, but
rather that it prefers to avoid only a certain class of subjects, namely those
which are new to the discourse.  Given that all of the examples above are
presented out of context, it is clear that all critical NPs are new to the
hearer/reader's model of the discourse.

\mysection{Given and New}%
Prince (1981)\nocite{Prince81} proposed a classification of occurrences of NPs
in terms of assumed familiarity.  When a speaker refers to an entity which
s/he assumes salient/familiar to the hearer, s/he tends to use a brief form,
such as a definite NP or a pronoun.  Otherwise s/he is obliged to provide the
hearer with enough information to construct this entity in the hearer's mind.
Prince classified the forms of NPs and ranks them from given to new:

\begin{description}
\item[evoked]  An expression used to refer to one of the conversation's
participants or
an entity which is already under discussion.  (usually a definite NP or
 pronoun)
\item[unused] A proper name which refers to an entity known to the speaker
and hearer, but
not already in the present discourse.
\item[inferable] A phrase which introduces an entity not already in the
discourse, but which is easily inferred from another entity currently under
discussion. (c.f.
bridging inference of Haviland and Clark 1974\nocite{HavilandClark74})
\item[containing inferable] An expression that introduces a new entity and
contains a
reference to the extant discourse entity from which the inference is to
proceed. (e.g. `one
of the people that work at Penn')
\item[brand new] An expression that introduces a new entity which cannot be
inferentially related or predicted from entities already in the discourse.
\end{description}
\noindent Prince constructed this scale on the basis of scale-based
implicatures that can be
drawn if a speaker uses a form which is either too high or too low ---
such a speaker would
be sounding uncooperative/cryptic or needlessly verbose, respectively.

Using this classification, Prince found that naturally occurring texts exhibit
a significant tendency to avoid placing new NPs (including inferable and
unused) in subject position.  If we construe this tendency as a principle of
the linguistic competence, we would indeed expect a reader to prefer to treat
an out-of-context NP as something other than a subject.  I refer to this
principle as {\em Avoid New Subjects.\/}

\mysection{Late Closure and Avoid New Subjects}%
The principle of Avoid New Subjects predicts that in ordinary text the Late
Closure Effect exhibited in \ref{LC1}b should disappear when the critical NP
is given in the discourse. To test this prediction, I conducted a survey of
the bracketed Brown and Wall Street Journal corpora for the following
configuration: a VP which ends with a verb and is immediately followed by an
NP.  Crucially, no punctuation was allowed between the VP and the NP.  I then
removed by hand all matches where there was no ambiguity, e.g. the clause was
in the passive or the verb could not take the NP as argument for some reason.
Of the eleven remaining matches, \ifshorten four \fi are are given
below.  Each is preceded by some context, and followed by illustration of the
ambiguity (in brackets), and by a categorization of the critical NP.
\begin{enumerate}
\item{} 
[An article about a movie describes how its composer approached one of the
singers.]  When you approach a singer and tell her you don't want her to sing
you always run the risk of offending.

[`You don't want her to sing you a song.']

`you' = evoked.
\ifshorten
 \item 
 From the way she sang in those early sessions, it seemed clear
 that Michelle (Pfeiffer) had
 been listening not to Ella but to Bob Dylan. ``There was a
 pronunciation and approach that
 seemed Dylan-influenced,'' recalled Ms.  Stevens.  Vowels were swallowed,
 word endings were
 given short or no shrift.  ``When we worked it almost became a joke with us
 that I was
 constantly reminding her to say the consonants as well as the vowels.''

 [`When we worked it out$\ldots$']

`it' = pleonastic.
\fi
\item 
After the 1987 crash, and as a result of the recommendations of many studies,
``circuit breakers'' were devised to allow market participants to regroup and
restore orderly market conditions.  It's doubtful, though, whether circuit
breakers do any real good.  In the additional time they provide even more
order imbalances might pile up, as would-be sellers finally get their broker
on the phone.

[Even though this example involves a wh-dependency, the fact remains that the
NP `even more order imbalances' could be initially construed as a dative, as
in `In the additional time they provide even the slowest of traders, problems
could$\ldots$']

`even more order imbalances' = brand new
\item{} 
[Story about the winning company in a competition for teenage-run businesses,
its president, Tim Larson, and the organizing entity, Junior Achievement.]
For winning Larson will receive a \$100 U.S. Savings Bond from the
Junior Achievement national organization.

[$\ldots$winning Larson over to their camp$\ldots$]

`Larson' = evoked
\end{enumerate}
As can be expected of carefully written prose, none of the matches posed any
reading difficulty.  Of the eleven critical NPs, four were pleonastic, five
were evoked, one was inferable and one was brand new.  Prince's givenness
scale does not include pleonastic NPs, since they do not refer.  For the
present purpose, it suffices to note that Avoid New Subjects does not rule out
pleonastics.  While the numbers here are too small for statistical
inference,\footnote{ Given the high frequency of given subjects, optionally
transitive verbs and fronted adverbials, one might expect more matches in a
two million word corpus.  But examination of the Wall Street Journal corpus
reveals that most fronted adverbials are set off by comma, regardless of
potential ambiguity.  Of 7256 sentence initial adverbials, only 8.14\% (591)
are not delimited by comma.  Of these 7256 adverbials, 1698 have the category
SBAR, of which only 4.18\% (71) are not delimited by comma.  The great
majority of fronted adverbials (4515) have category PP, of which 8.75\% (433)
are not delimited by comma.  } the data suggest that the prediction of Avoid
New Subjects is maintained.

\section{Complement Clauses}

\vspace{-2.4mm}

\noindent%
In order to be relevant for the ambiguity in \ref{MA2} Avoid New Subjects must
be applicable not just to subjects of matrix clauses but also to embedded
subjects.  It is widely believed that constituents in a sentence tend to be
ordered from given to new.  The statistical tendency to avoid new subjects may
be arising solely as a consequence of the tendency to place new information
toward the end of a sentence and the grammatically-imposed early placement of
subjects.  If this were the case, that is, Avoid New Subjects is a corollary
of Given Before New, then Avoid New Subjects would make no predictions about
subjects of complement clauses, given that complement clauses tend to appear
rather late in the sentence.  I now argue from the perspective of sentence
production that it is the grammatical function of subjects, not just their
linear placement in the sentence, that is involved with the avoidance of new
information.

When a speaker/writer wishes to express a proposition which involves reference
to an entity not already mentioned in the discourse, s/he must use a {\em
new\/} NP. S/he is quite likely to avoid placing this NP in subject position.
To this end, s/he may use constructions such as passivization,
there-insertion, and clefts.

Avoid New Subjects predicts that this sort of effort on behalf of writers
should be evident in both matrix clauses and complement clauses.  To test this
prediction, I compared the informational status of NPs in subject and
non-subject positions in both matrix and embedded clauses.  I defined subject
position as an NP immediately dominated by S and followed (not necessarily
immediately, to allow for auxiliaries, punctuation, etc.) by a VP.  I defined
non-subject position as an NP either immediately dominated by VP or
immediately dominated by S and not followed (not necessarily immediately) by
VP.  To determine givenness status, I used a simple heuristic
procedure\footnote{I am grateful to Robert Frank for helpful suggestions
regarding this procedure.} to classify an NP into one of the following
categories: {\sc empty-category}, {\sc pronoun}, {\sc proper-name}, {\sc
definite}, {\sc indefinite}, {\sc not-classified}.
\ifshorten
 The observed frequencies are given in table 1 at the end of this paper.
\fi

{\sc pronoun}s are either pleonastic or evoked --- they are thus fairly
reliable indicators of given (at least non-new) NPs.  The category {\sc
indefinite} contains largely brand-new or inferable NPs, thus being a good
indicator of new information.  Considering {\sc pronoun}s and {\sc
indefinite}s there is a clear effect on grammatical function for both matrix
and embedded clauses.%
\ifshorten%
\footnote{For reasons
 of space I only give results from the Brown corpus, but all assertions I make
 also hold of
 the Wall Street Journal corpus.}
\else
\fi

\mytable{
\begin{tabular}{|l|rr|rr|}
\hline
 & \multicolumn{2}{c|}{matrix clause} & \multicolumn{2}{c|}{embedded clause}\\
		  &  subj  & non-subj &  		subj & non-subj\\
	\hline
{\sc pronoun}		  &   7580 &   956    &  		1800 &   213 \\
{\sc indefinite}	  &   4157 &  5269    &  		 736 &   899\\
 \hline
                  & \multicolumn{2}{c|}{$\chi^2 = 3952.2$}
                       & \multicolumn{2}{c|}{$\chi^2 = 839.5$} \\
                  & \multicolumn{2}{c|}{$p<0.001$}
                       & \multicolumn{2}{c|}{$p<0.001$} \\ \hline
\end{tabular}}

\noindent The prediction of Avoid New Subjects is therefore verified.

When a hearer/reader is faced with an initial-segment such as
\ref{HeardAgain}, the ambiguity is not exactly between an NP complement
analysis versus an S-complement analysis, but rather between an TR (transitive
verb) analysis and an RC (reduced S-complement).

\startxl{HeardAgain}
John has heard the joke$\ldots$
\stopx

It is therefore necessary to verify that Avoid New Subjects is indeed
operating in this RC sub-class of sentential complements.  A further analysis
reveals that this is indeed the case.

\mytable{
\begin{tabular}{|l|rr|rr|}
\hline
 & \multicolumn{2}{c|}{TC} &  \multicolumn{2}{c|}{RC}\\
		  &  subj  & non-subj &  subj & non-subj\\ \hline
{\sc pronoun}		  &    773 &    79   &   1027 &    134\\
{\sc indefinite}	  &    617 &   555   &    119 &    344\\ \hline
                  & \multicolumn{2}{c|}{$\chi^2 = 332.6$}
                            & \multicolumn{2}{c|}{$\chi^2 = 627.6$}  \\
                  & \multicolumn{2}{c|}{$p<0.001$}
                            & \multicolumn{2}{c|}{$p<0.001$}  \\ \hline
\end{tabular}}

\noindent If anything,  Avoid New Subjects has a stronger effect after a zero
complementizer.

\mysection{Unambiguous Structures}%
The findings of Holmes \etal\ (1987), that effects predicted by Avoid New
Subjects are present even in unambiguous structures, transfer to other
unambiguous structures, such as the classical center embedding sentence:

\startxl{ratcat}
The rat that the cat that the dog bit chased died.
\stopx

\noindent In this sentence three new subjects must be accommodated
simultaneously.  Of all the examples that I have seen of sentences with a
structure as in \ref{ratcat}, the easiest one to understand is \ref{Italian}
from Frank (1992)\nocite{Frank92}.

\startxl{Italian}
A book that some Italian I've never heard of wrote will be published soon by
MIT press.
\stopx

\noindent Notice that this example requires the simultaneous processing of
only two new subjects, the most embedded subject being evoked.  When this
subject is replaced with a new NP, the sentence becomes harder to process.

\startxl{Italian2}
A book that some Italian most people have never heard of wrote will be
published soon by MIT press.
\stopx

Also, when a complex, center-embedded NP appears in subject position
\ref{bureaucrats}a, it is harder to process than when it appears in object
position \ref{bureaucrats}b. \ref{bureaucrats} is based on an example from
Gibson (1991)\nocite{Gibson91} which is in turn based on earlier work by
Cowper.

\startxl{bureaucrats}
\myitem{a.}{Many bureaucrats
	who the information that Iraq invaded Kuwait affected negatively
work for the government.}
\myitem{b.}{The government employs many bureaucrats who the information that
  Iraq invaded Kuwait affected negatively.}
\stopx

I don't wish to claim that the new-subject effect is solely responsible for
all of the difficulty associated with center embedding, but it is clear that
it is playing an important role.

\vspace{-4mm}

\mysection{Conclusion}%
I have argued for an account of sentence processing wherein the syntactic
processor applies the rules of the competence grammar blindly and faithfully.
Ambiguity resolution decision are made by the interpreter when it considers
the analyses which the syntactic processor has proposed and evaluates them on
the basis of sensibleness.  The criteria I have appealed to: Grice's maxims,
ordering constituents by the amount of information they convey, and not
putting new information in subject position, are all components of our
knowledge of language and not the exclusive domain of the process of parsing,
nor that of production.

Avoid New Subject predicts that for sentences such as \ref{MA2}, the garden
path effect should disappear when the sentence is put in a context in which
the NP is given, and its form is felicitous.  The design and execution of an
experiment to test this prediction remain for future research.

\begin{table*}
\vbox{
\noindent Brown Corpus:

\centerline{\begin{tabular}{|l|rrrr|rrrr|}  \hline
                 & \multicolumn{4}{c|}{Subjects} & \multicolumn{4}{c|}
	{Non Subjects} \\ \hline
      givenness status &{\sc tc}&{\sc rc}&{\sc tc+rc}&matrix&{\sc tc}&
	{\sc rc}&{\sc tc+rc}
							  &matrix\\ \hline
  {\sc empty-category}
	&     0 &    50 &    50 &     0  &     6 &    41 &    47 &     0 \\
         {\sc pronoun}
	&   773 &  1027 &  1800 &  7580  &    79 &   134 &   213 &   956 \\
     {\sc proper-name}
	&   201 &    81 &   282 &  2838  &    32 &    21 &    53 &   539 \\
        {\sc definite}
	&   890 &   266 &  1156 &  6686  &   351 &   182 &   533 &  3399 \\
      {\sc indefinite}
	&   617 &   119 &   736 &  4157  &   555 &   344 &   899 &  5269 \\
  {\sc not-classified}
	&   259 &   107 &   366 &  3301  &   167 &    79 &   246 &  1516 \\
\hline    total:
	&  2740 &  1650 &  4390 & 24562  &  1190 &   801 &  1991 &  11679 \\
 \hline
\end{tabular}}}
\vspace{1ex}

\vbox{
\noindent Wall Street Journal Corpus:

\centerline{\begin{tabular}{|l|rrrr|rrrr|}  \hline
                 & \multicolumn{4}{c|}{Subjects} &
		\multicolumn{4}{c|}{Non Subjects} \\ \hline
      givenness status &{\sc tc}&{\sc rc}&{\sc tc+rc}&matrix&{\sc tc}&{\sc
	rc}&{\sc tc+rc}
							   &matrix\\ \hline
  {\sc empty-category}
	&     4 &    83 &    87 &     0  &     2 &    20 &    22 &     1 \\
         {\sc pronoun}
	&   369 &  2263 &  2632 &  2347  &    34 &    90 &   124 &   169 \\
     {\sc proper-name}
	&   167 &   371 &   538 &  3364  &    29 &    89 &   118 &   377 \\
        {\sc definite}
	&   610 &  1686 &  2296 &  5385  &   253 &   729 &   982 &  1959 \\
      {\sc indefinite}
	&   498 &   805 &  1303 &  3847  &   484 &  1375 &  1859 &  4039 \\
  {\sc not-classified}
	&   251 &   278 &   529 &  2402  &   178 &   581 &   759 &  2138 \\
\hline    total:
	&  1899 &  5486 &  7385 & 17345  &   980 &  2884 &  3864 &   8683 \\
\hline
\end{tabular}}}
\caption{Frequencies of NPs by discourse status and grammatical position in
two corpora.}
\end{table*}

\setlength{\parindent}{-3mm}
\addtolength{\hsize}{-1mm}
\addtolength{\columnwidth}{-1mm}
\setlength{\parskip}{.8mm}

\section*{Bibliography}

\vspace{-5mm}

\rule{0em}{0em}

\indent Altmann, Gerry and Mark J. Steedman.~1988.
Interaction with Context during Human Sentence
Processing.  {\em Cognition~30.}

Crain, Stephen and Mark J. Steedman.~1985. On not being led up the garden
path.  In David Dowty \etal\ eds.  {\em Natural Language Parsing:
Psychological Computational and Theoretical Perspectives.} Cambridge.

Fodor, Janet D. and Lyn Frazier.~1978.  The Sausage Machine: a new two-stage
parsing model.  {\em Cognition~6.}

Frank, Robert E.~1992.  {\em Syntactic Locality and Tree Adjoining Grammar:
Grammatical, Acquisition and Processing Perspectives.}
Ph.D. Dissertation. University of Pennsylvania

Frazier, Lyn and Keith Rayner.~1982.  Making and Correcting Errors During
Sentence Comprehension: Eye Movements in the Analysis of Structurally
Ambiguous Sentences. {\em Cognitive Psychology~14.}

Garnsey, Susan M., Melanie A. Lotocky and George W. McConkie.~1992.
Verb-Usage Knowledge in Sentence Comprehension.  Presented at the Psychonomic
Society Meeting.~1992.

Gibson, Edward A.~F.~1991. {\em A Computational Theory of Human Linguistic
Processing: Memory Limitations and Processing Breakdown} Ph.D. Dissertation.
Carnegie Mellon University.

Grice, H. Paul.~1975. Logic and Conversation. In Peter Cole and Jerry
L. Morgan, eds.  {\em Syntax and Semantics~3. Speech Acts.} New York. Academic
Press.

Haviland, Susan and Herbert Clark.~1974. What's new? Acquiring New Information
as a Process in Comprehension. {\em Journal of Verbal Learning and Verbal
Behavior~13.}

Holmes, Virgina M., Alan Kennedy and Wayne S. Murray.~1987. Syntactic
Structure and the Garden Path. {\em The Quarterly Journal of Experimental
Psychology~39A}.

Kennedy, Alan, Wayne S. Murray, Francis Jennings and Claire
Reid.~1989. Parsing Complements: Comments on the Generality of the Principle
of Minimal Attachment. {\em Language and Cognitive Processes~4}.

Ladd, D. Robert.~1992.  {\em Compound Prosodic Domains.} University of
Edinburgh Occasional Paper.

Mitchell, Don C., Martin M. B. Corley and Alan Garnham.~1992.  Effects of
Context in Human Sentence Parsing: Evidence Against a Discourse-Based Proposal
Mechanism. {\em Journal of Experimental Psychology: Learning, Memory and
Cognition~18}

Niv, Michael.~1992.  Right Association Revisited.  In {\em Proceedings of the
30\th{th} Annual Meeting of the Association for Computational Linguistics.}

Pearlmutter, Neal J. and Maryellen C. MacDonald.~1992.  Plausibility and
syntactic ambiguity resolution.  In {\em Proceedings of the 14\th{th} Annual
Conference of the Cognitive Science Society.\/} Hillsdale, New
Jersey. Erlbaum.

Prince, Ellen F.~1981. Toward a Taxonomy of Given/New Information. In Peter
Cole, ed. {\em Radical pragmatics.} New York. Academic Press.

Pritchett, Bradley L.~1988. Garden Path Phenomena and the Grammatical Basis of
Language Processing. {\em Language~64.}

Stowe, Laurie A.~1989. Thematic Structures and Sentence Comprehension. In Greg
N. Carlson and Michael K. Tanenhaus, eds. {\em Linguistic Structure in
Language Processing.} Kluwer Academic Publishers.

Trueswell, John C. and Michael K. Tanenhaus.~1991. Tense, Temporal Context and
Syntactic Ambiguity Resolution. {\em Language and Cognitive Processes~6}.

\end{document}